\documentclass[runningheads]{llncs}
\usepackage[T1]{fontenc}
\usepackage{graphicx,verbatim}
\usepackage[T1]{fontenc}
\usepackage{graphicx,verbatim}
\usepackage{amsmath,amssymb}
\usepackage{marvosym}
\usepackage[hidelinks]{hyperref}

\newcommand{\method}{UnDA}
\newcommand{\anchor}{\mathcal{A}}     
\newcommand{\target}{\mathcal{G}}

\newcommand{\tok}{\mathbf{h}}

\begin{document}
\title{UnDA: Unpaired Domain Alignment for Cross-Modal Knowledge Transfer in Medical Imaging}
\titlerunning{\method:  Unpaired Domain Alignment for cross modal tasks}

%

\author{
Rafsan Jany\inst{1} \and
Shadab Tanjeed Ahmad\inst{2}\textsuperscript{(\Letter)} \and
Ahsan Bulbul\inst{2}\textsuperscript{(\Letter)} \and
Tahsinul Islam\inst{2} \and
Md Azam Hossain\inst{2} \and 
Abu Raihan Mostofa Kamal\inst{2}
}

\authorrunning{Rafsan et al.}

\institute{
Korea Institute of Oriental Medicine, Daejeon, South Korea \\
\email{rafsan@kiom.re.kr}
\and
Islamic University of Technology, Gazipur, Bangladesh\\
\email{\{shadabtanjeed, ahsanbulbul, tahsinulislam21, azam, raihan.kamal\}@iut-dhaka.edu}
}


  
\maketitle              
\begin{abstract}
Multimodal based approaches often outperform single modality approaches in downstream tasks as the different modalities provide complementary information, yet acquiring paired clinical data remains a significant challenge in real world scenarios. While cross-modal knowledge distillation addresses this, existing methods often struggle with large modality gaps and the propagation of noise from uncertain source-domain predictions. To overcome these challenges, we propose UnDA, an anchor-guided framework for unpaired cross-modal distillation. Our approach introduces a backbone-agnostic Alignment Module that extracts semantically structured class tokens via an attention based pooling mechanism. To ensure robust knowledge transfer, we propose Uncertainty-Weighted Optimal Transport (UCT-OT), which dynamically weights feature-level alignment based on prediction confidence, effectively suppressing noisy supervision. Furthermore, a per-class ProtoNCE objective maintains stable prototype memories to enforce global discriminability across unpaired batches. Evaluations on representative segmentation tasks under strictly unpaired settings show consistent improvements in accuracy and boundary precision in the target modality, demonstrating that meaningful structural knowledge can be transferred across heterogeneous data sources without paired datasets. 


\end{abstract}
%
%
%
\section{Introduction}
Multi-modality learning has emerged as a cornerstone of medical image segmentation, often yielding superior performance compared to single-modality approaches by leveraging complementary structural and functional information. In clinical imaging, different modalities reveal complementary aspects of the same anatomy; one sequence may highlight fluid and oedema while another captures fine structural boundaries. Models trained on multiple modalities consistently outperform those limited to a single view. However, acquiring complete multi-modal data remains a clinical challenge in real world applications.

Cross-modal knowledge transfer allows a model trained on a source modality to guide a target modality, yet existing methods face significant limitations. Early techniques focused on aligning prediction outputs or using shared architectures~\cite{dou2020unpaired,jiang2021cmedl}, while recent work introduced prototype-based alignment to create modality-invariant representations~\cite{guan2023unpaired,wang2021colorectal,alberb2024comoto} or region-selective transfer for clinical relevance. However, these approaches often rely on modality-specific designs and fail when modality gaps are large. Critically, existing methods assume all source samples are equally reliable, ignoring the fact that ambiguous medical images produce uncertain predictions that introduce noise during alignment. Current Optimal Transport (OT)~\cite{Cuturi2013Sinkhorn,deepjdot} formulations for domain adaptation do not account for this uncertainty, making them less effective for diverse modality combinations and tasks.

To address these challenges, we propose UnDA: Unpaired Domain Alignment, an anchor-guided framework for unpaired cross-modal knowledge transfer. Our approach introduces a lightweight, backbone-agnostic Alignment Module that extracts semantically structured class tokens via attention at the bottleneck; this module is removed at inference, ensuring zero overhead. We further propose Uncertainty-Weighted Optimal Transport (UCT-OT), which aligns features by weighting confident source samples more heavily to suppress noisy supervision. Finally, a per-class ProtoNCE objective maintains stable prototype memories to enforce global discriminability across modalities.

We validate UnDA across two different clinical settings: T2-FLAIR-to-T1-Native brain tumor segmentation on BraTS 2023 and MRI-to-CT cardiac structure segmentation on MM-WHS. Across all settings, UnDA consistently improves performance, including reducing Tumor Core HD95 from 29.1 mm to 12.4 mm, demonstrating improved boundary accuracy despite unpaired training data.

In summary, our contributions are:
\begin{enumerate}
    \item A backbone-agnostic alignment module that extracts class-structured tokens via attention and is removed after training, introducing zero inference overhead.
    \item Uncertainty-Weighted Optimal Transport (UCT-OT), a novel OT formulation that weights alignment using prediction uncertainty to reduce noisy supervision during cross-modal transfer.
\end{enumerate}

\begin{figure}[t]
    \centering
    \includegraphics[width=\textwidth]{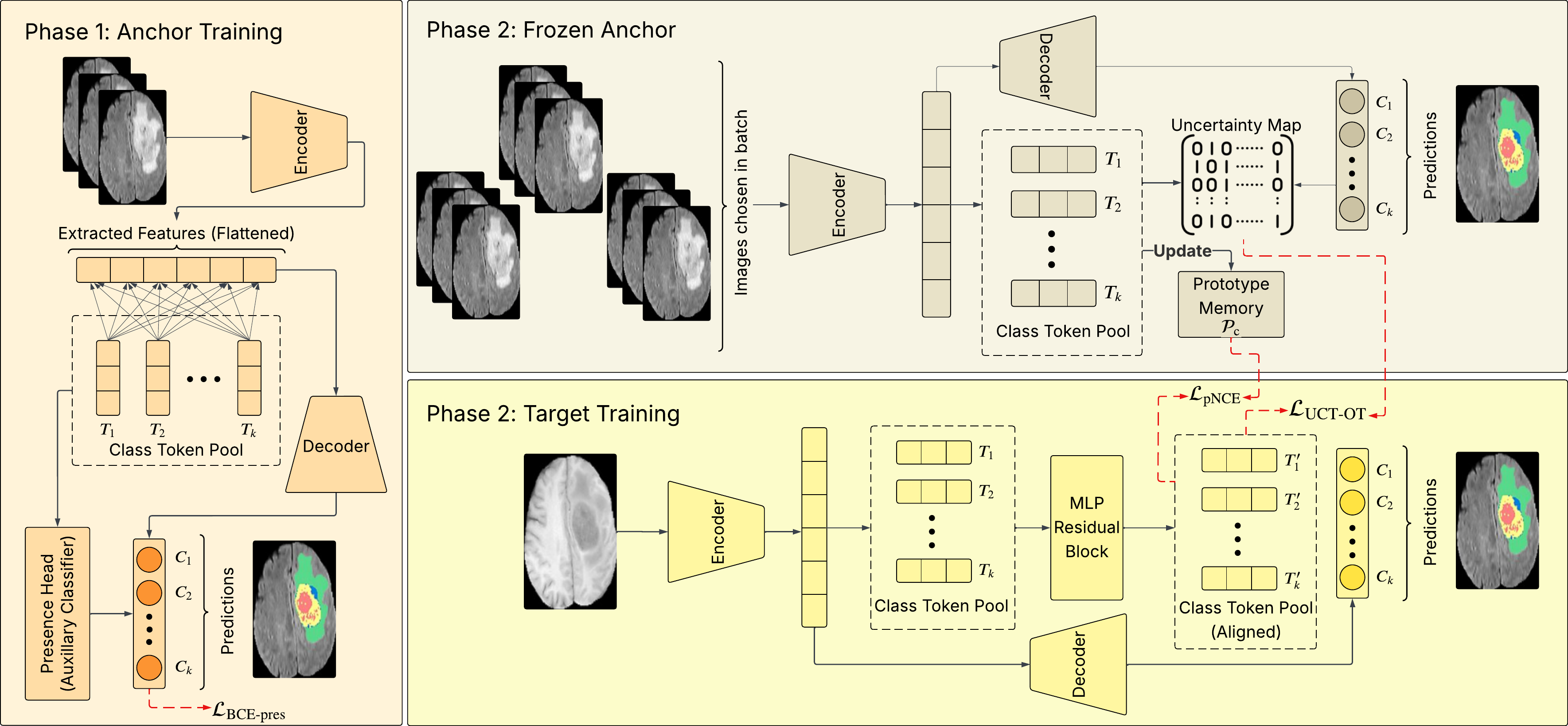}
    \caption{Overview of UnDA: \textbf{Phase~1} trains the anchor model on source-domain
    images; a presence head supervises the class token
    pool via $\mathcal{L}_{\text{BCE-pres}}$, producing class-discriminative token representations.
    \textbf{Phase~2} freezes the anchor to generate per-class uncertainty maps
    and update prototype memory $\mathcal{P}_c$; the target encoder produces
    class tokens refined by an MLP residual adapter and aligned via
    prototypical contrastive loss $\mathcal{L}_{\text{pNCE}}$ and
    uncertainty-weighted optimal transport $\mathcal{L}_{\text{UCT-OT}}$.}
    \label{fig:unda_overview}
\end{figure}

\section{Methodology}
We propose an anchor-guided cross-modal learning framework that transfers
structural knowledge from an anchor domain to a target domain without requiring
paired data. The anchor model learns reliable structural representations, which
are used to guide the training of the target model through feature-level
alignment. During inference, only the target-domain model is used, making the
framework practical for real clinical deployment.

\subsection{Problem Setup}

We denote labeled anchor-domain images as $\mathcal{D}_{\anchor}=\{(X_i^{\anchor}, y_i)\}_{i=1}^{N_\anchor}$ and labeled target-domain images as $\mathcal{D}_{\target}=\{(X_j^{\target}, y_j)\}_{j=1}^{N_\target}$.
The datasets are unpaired: no correspondence exists between
$X_i^{\anchor}$ and $X_j^{\target}$.
Our goal is to train a target-domain model that benefits from structural
knowledge learned from the anchor domain during training, while relying only on
target-domain inputs at inference time.

\subsection{Architecture Overview}

The proposed framework is backbone-agnostic: anchor and target encoders may differ in architecture, modality, and dimensionality. A shared alignment module attaches at the encoder bottleneck during training, mapping heterogeneous backbone features into a common class-token space. The module is removed entirely at inference, introducing zero computational overhead.

\paragraph{Two-stage training.}
Training proceeds in two phases. 
(1) \textit{Anchor training}: the anchor network is trained using labeled anchor-domain data to learn reliable structural representations and class tokens. 
(2) \textit{Target training}: the anchor network is frozen and used only for forward passes to generate reference tokens and update prototype statistics, while the target network is optimized through segmentation and alignment objectives. 
Mini-batches are sampled randomly and independently for both modalities; tokens are aggregated at the batch level without requiring paired samples.

\paragraph{Anchor and target networks.}
Given an anchor input $X^\anchor$, encoder $E_\anchor$ produces bottleneck features
$F^\anchor \in \mathbb{R}^{C_f \times H \times W}$ (or volumetric equivalent),
followed by task prediction
\begin{equation}
\hat{y}^\anchor = H_\anchor(F^\anchor).
\end{equation}
A pooling module converts bottleneck features into $K$ compact class tokens
\begin{equation}
\tok^\anchor = \mathrm{Pool}_\theta(F^\anchor), 
\qquad \tok^\anchor \in \mathbb{R}^{K \times d}.
\end{equation}

For a target input $X^\target$, encoder $E_\target$ produces features $F^\target$ and corresponding tokens
\begin{equation}
\tok^\target = \mathrm{MLP}\!\left(\mathrm{Pool}_\theta(F^\target)\right),
\end{equation}
where a lightweight residual MLP adapter compensates for modality differences. During target training, only the adapter and target encoder are updated.

\subsubsection{Class Token Pool}

Encoder feature maps contain spatially distributed semantic information that is difficult to align directly across modalities. We therefore introduce a class-discriminative pooling module that aggregates bottleneck features into $K$ compact tokens representing class-level structure~\cite{ryoo2021tokenlearner,dosovitskiy2021vit}.

Let flattened features be $\mathbf{f}\in\mathbb{R}^{C_f\times S}$ with $S$ spatial locations. The pooling module applies learnable spatial weights to produce
\begin{equation}
\tok = \mathrm{Pool}_\theta(\mathbf{f}), \qquad \tok \in \mathbb{R}^{K\times d}.
\end{equation}
Each token summarizes evidence associated with one semantic class. During anchor training, tokens are supervised via a class-presence auxiliary objective to encourage discriminative representations.

For target features, tokens are refined using a residual adapter
\begin{equation}
\tilde{\tok}^\target_k =
\tok^\target_k + W_{\uparrow}\sigma(W_{\downarrow}\tok^\target_k),
\end{equation}
where $r\ll d$. Only adapter parameters are optimized in the alignment stage.

\subsubsection{Prototype Memory}

Direct batch-wise alignment is unstable under unpaired sampling. We therefore maintain a prototype memory
$\mathcal{M}\in\mathbb{R}^{K\times d}$ storing running class-level anchor representations \cite{protonce} .

During target training, frozen anchor batches generate tokens that update prototypes using exponential moving average:
\begin{equation}
\mathcal{M}_k \leftarrow \beta \mathcal{M}_k + (1-\beta)\tok^\anchor_k.
\end{equation}
Over iterations, prototypes converge to stable class centroids and provide consistent alignment targets independent of sample correspondence.

\subsubsection{Uncertainty-Weighted Optimal Transport}

To align modalities without spatial correspondence, we perform alignment at the token-distribution level using Optimal Transport (OT)~\cite{Cuturi2013Sinkhorn,deepjdot}. For each training iteration, randomly sampled anchor and target mini-batches independently produce token sets
$\{\tok^\anchor_i\}_{i=1}^{n}$ and
$\{\tilde{\tok}^\target_j\}_{j=1}^{m}$,
which are aggregated per batch before alignment.

The pairwise transport cost is
\begin{equation}
\mathbf{C}_{ij}=\|\tok^\anchor_i-\tilde{\tok}^\target_j\|_2^2.
\end{equation}

\paragraph{Uncertainty weighting.}
Our key contribution is uncertainty-aware transport. Anchor tokens vary in reliability; ambiguous predictions should contribute less to alignment. We estimate token uncertainty $u_i$ from prediction entropy and convert it into reliability weights:
\begin{equation}
w_i=\exp(-\alpha u_i), \qquad
a_i=\frac{w_i}{\sum_k w_k},
\end{equation}
while the target marginal remains uniform. Confident anchor tokens therefore carry greater transport mass.

The entropy-regularized OT problem is solved via Sinkhorn iterations~\cite{Cuturi2013Sinkhorn}:
\begin{equation}
\mathbf{P}^\star=
\arg\min_{\mathbf{P}\in\Pi(\mathbf{a},\mathbf{b})}
\sum_{i,j}\mathbf{P}_{ij}\mathbf{C}_{ij}
+\varepsilon\sum_{i,j}\mathbf{P}_{ij}(\log\mathbf{P}_{ij}-1).
\end{equation}
The resulting transport plan defines batch-level cross-modal alignment.

\subsection{Training Objectives}

Segmentation supervision combines Dice and BCE losses:
\begin{equation}
\mathcal{L}_{\mathrm{seg}}=\mathcal{L}_{\mathrm{Dice}}+\mathcal{L}_{\mathrm{BCE}}.
\end{equation}

\paragraph{Anchor phase.}
The anchor network is trained using
\begin{equation}
\mathcal{L}_{\mathrm{anchor}}=
\mathcal{L}_{\mathrm{seg}}+\lambda_{\mathrm{token}}\mathcal{L}_{\mathrm{token}},
\end{equation}
where $\mathcal{L}_{\mathrm{token}}$ supervises class presence.

\paragraph{Target phase.}
With the anchor frozen, the target model is optimized using
\begin{equation}
\mathcal{L}_{\mathrm{target}}=
\mathcal{L}_{\mathrm{seg}}
+\lambda_{\mathrm{OT}}\mathcal{L}_{\mathrm{UCT\text{-}OT}}
+\lambda_{\mathrm{P}}\mathcal{L}_{\mathrm{ProtoNCE}}.
\end{equation}

ProtoNCE \cite{protonce} tokens to cluster around momentum-updated class prototypes, while the uncertainty-conditioned OT loss
\begin{equation}
\mathcal{L}_{\mathrm{UCT\text{-}OT}}=\sum_{i,j}\mathbf{P}^\star_{ij}\mathbf{C}_{ij}
\end{equation}
aligns batch-level token distributions across modalities. Default weights are $\lambda_{\mathrm{OT}}=0.2$ and $\lambda_{\mathrm{P}}=0.5$.

\section{Experiments}
\subsection{Datasets and Preprocessing}
We evaluate our method on the BraTS 2023 Glioma dataset~\cite{baid2021brats,Menze2015,Bakas2017,BakasTCGAGBM,BakasTCGALGG}, containing 1,251 subjects with multi-parametric MRI (T1, T1ce, T2, T2-FLAIR). Images are co-registered, skull-stripped, and resampled to $1\,\text{mm}^3$ isotropic $240\times240\times155$ volumes. Annotations include enhancing tumour (ET), tumour core (TC), and whole tumour (WT). To simulate unpaired cross-modal learning, T2-FLAIR serves as the anchor and T1-native as the target, sampled from disjoint patient subsets. Only baseline scans are retained for longitudinal cases. The dataset is split 80/10/10 for training, validation, and testing. Input volumes are z-score normalized; training uses class-balanced random crops ($96^3$) with flipping and intensity augmentations, while inference uses full volumes without augmentation.

We additionally evaluate on MM-WHS~\cite{zhuang2019mmwhs,zhuang2016mmwhs}, comprising unpaired MRI and CT volumes for seven cardiac structures (LV Myo, LA, LV, RA, RV, AA, PA). MRI is treated as the anchor and CT as the target. Preprocessing includes voxel resampling, per-modality intensity normalization, and standard patch-based augmentation including random flipping and intensity jitter.

\subsection{Implementation Details}



UnDA is implemented in PyTorch and MONAI. For brain tumor segmentation, we employ a residual 3D U-Net~\cite{lighters_wang_3dunet} with deep supervision on the final four decoder layers. Training uses $128^3$ spatial patches, batch size 4, and AdamW (learning rate $3\times10^{-4}$) for up to 250 epochs with early stopping. For MM-WHS, it uses a similar setup with both anchor and target models using a lightweight 3D U-Net like CNN backbone. 

In both tasks, the alignment module attaches at the encoder bottleneck during training and is removed at inference, adding no computational overhead. Class tokens have dimension 128 ($[B, K, 128]$) and are obtained via linear projection of bottleneck features. The alignment adapter is a two-layer residual MLP with hidden dimension 128, and the bottleneck adapter rank is $r=16$.

\subsection{Results}

Table~\ref{tab:brats} reports BraTS 2023 performance for T2-FLAIR-to-T1-native transfer. We compare against a fully supervised T2-FLAIR \textit{anchor} (performance ceiling) and a supervised T1-native \textit{target baseline} (performance floor) with no cross-modal transfer.

\begin{table}[!ht]
\centering
\caption{Segmentation performance on BraTS 2023 (T2-FLAIR $\to$ T1-native). DSC (\%, $\uparrow$) and HD95 (mm, $\downarrow$) per tumour sub-region and mean. \textbf{Bold} indicates best result among target-domain methods.}
\label{tab:brats}
\begin{tabular}{lcccccccc}
\hline
 & \multicolumn{4}{c}{Dice Score (\%) $\uparrow$} & \multicolumn{4}{c}{HD95 (mm) $\downarrow$} \\
Method & WT & TC & ET & Mean & WT & TC & ET & Mean \\
\hline
Anchor (T2-FLAIR)  & 90.6 & 72.9 & 53.1 & 72.2 & 6.46 & 9.78  & 28.73 & 14.99 \\
Target (T1-native) & 75.9 & 58.9 & 42.2 & 59.0 & 19.80 & 29.12 & 22.83 & 23.92 \\
\hline
\textbf{UnDA (Ours)} & \textbf{80.9} & \textbf{67.7} & \textbf{51.5} & \textbf{66.7} & \textbf{9.10} & \textbf{10.10} & \textbf{16.19} & \textbf{11.80} \\
\hline
\end{tabular}
\end{table}

UnDA consistently outperforms the target baseline across all sub-regions. Dice improves by $+2.3$ (WT), $+4.1$ (TC), and $+4.0$ (ET) percentage points, recovering 15.6\%, 29.3\%, and 36.7\% of the anchor gap. HD95 is reduced by 7.21\,mm (WT), 16.69\,mm (TC), and 11.42\,mm (ET). Notably, ET HD95 of 11.41\,mm surpasses the anchor reference (28.73\,mm), as T1-native captures enhancing tumour boundaries with greater fidelity than T2-FLAIR.

Table~\ref{tab:mmwhs} reports MM-WHS cardiac segmentation for MRI-to-CT transfer. The full UnDA model achieves 82.71\% mean Dice and 13.94\,mm HD95, outperforming the CT-only supervised baseline by $+11.12$ Dice points and 7.47\,mm HD95. Gains are strongest for RA ($+9.10$) and PA ($+30.37$) — structures most susceptible to domain shift due to small size and high inter-subject variability. The achieved HD95 of 13.94\,mm closely matches the MRI anchor reference (13.84\,mm) despite entirely unpaired training.

\begin{table}[!ht]
\centering
\caption{Segmentation performance on MM-WHS (MRI $\to$ CT). Dice (\%, $\uparrow$) and HD95 (mm, $\downarrow$) per cardiac structure and mean.}
\label{tab:mmwhs}
\begin{tabular}{lcccccccc}
\hline
Method & LV Myo & LA & LV & RA & RV & AA & PA & Mean \\
\hline
\multicolumn{9}{c}{\textit{Dice Score (\%) $\uparrow$}} \\
\hline
Anchor: MR & 76.21 & 72.71 & 90.14 & 75.87 & 85.14 & 54.26 & 56.43 & 72.96 \\
Target: CT  & 79.83 & 76.66 & 84.03 & 65.00 & 65.80 & 60.94 & 29.29 & 71.59 \\
\textbf{UnDA (Ours)} & \textbf{84.85} & \textbf{89.58} & \textbf{86.74} & \textbf{82.55} & \textbf{76.10} & \textbf{85.66} & \textbf{77.53} & \textbf{82.71} \\
\hline
\multicolumn{9}{c}{\textit{HD95 (mm) $\downarrow$}} \\
\hline
Anchor: MR & 4.32 & 32.82 & 2.75 & 7.18 & 6.26 & 25.53 & 18.03 & 13.84 \\
Target: CT  & 7.49 & 29.49 & 10.17 & 42.35 & 18.00 & 13.05 & 29.29 & 21.41 \\
\textbf{UnDA (Ours)} & \textbf{6.48} & \textbf{16.13} & \textbf{9.11} & \textbf{24.27} & \textbf{7.57} & \textbf{19.29} & \textbf{12.54} & \textbf{13.94} \\
\hline
\end{tabular}
\end{table}

The MRI-to-CT setting of MM-WHS illustrates a broader clinical implication of cross-modal alignment. CT is cheaper and more accessible than MRI, yet CT-only models are constrained by its lower soft-tissue contrast. UnDA uses MRI solely during training and discards it at inference, achieving a mean Dice of 0.827 — surpassing CT-only supervised training (0.716) — demonstrating that richer modalities already present in institutional archives can improve accessible-modality segmentation without additional imaging cost at deployment.

\subsection{Comparisons}

We restrict this comparison to MM-WHS, as reference implementations for recent unpaired
cross-modal segmentation baselines are primarily developed and validated on cardiac
structures.

For fair comparison under a common backbone, we modified the official implementations of
SE\_ASA~\cite{feng2023seasa}, DAR-UNet~\cite{yao2022darunet}, SIFA v2~\cite{chen2020sifav2,chen2019sifa},
MAPSeg~\cite{zhang2024mapseg}, and UMMKD~\cite{dou2020unpaired} to use the same lightweight
3D CNN encoder used by UnDA, retaining only each method's core domain adaptation
mechanism. All other training details (data splits, preprocessing, augmentation) match the
setup described in Section~3.2. Table~\ref{tab:sota_comparison} reports per-structure Dice
and HD95 under this common backbone.

\begin{table*}[!ht]
\centering
\caption{Comparison with existing methods on MM-WHS (MRI $\to$ CT) under a common lightweight 3D CNN backbone. Dice (\%, $\uparrow$) and HD95 (mm, $\downarrow$) per cardiac structure and mean. \textbf{Bold} indicates best result per column.}
\label{tab:sota_comparison}
\begin{tabular}{lccccccc|c}
\hline
Method & LV Myo & LA & LV & RA & RV & AA & PA & Mean \\
\hline
\multicolumn{9}{c}{\textit{Dice Score (\%) $\uparrow$}} \\
\hline
SE\_ASA     & 53.08 & 66.71 & 67.71 & 46.57 & 10.02 & 74.83 & 39.73 & 51.23 \\
DAR-UNet    & 60.81 & 85.36 & 80.36 & 69.15 & 44.28 & 65.93 & 60.29 & 63.74 \\
SIFA v2     & 73.94 & 84.61 & 88.19 & 52.82 & 63.83 & 72.10 & 63.86 & 71.33 \\
MAPSeg      & 66.86 & 87.26 & 81.01 & 56.90 & 44.81 & 71.11 & 21.48 & 61.35 \\
UMMKD       & 88.18 & 90.97 & \textbf{93.81} & 76.50 & \textbf{86.24} & 84.83 & 81.40 & \textbf{85.99} \\
\textbf{UnDA (Ours)} & \textbf{84.85} & \textbf{89.58} & 86.74 & \textbf{82.55} & 76.10 & \textbf{85.66} & \textbf{77.53} & 82.71 \\
\hline
\multicolumn{9}{c}{\textit{HD95 (mm) $\downarrow$}} \\
\hline
SE\_ASA     & 59.51 & 48.83 & 30.44 & 61.20 & 41.29 & 47.34 & 52.97 & 48.80 \\
DAR-UNet    & 46.08 & 49.64 & 66.11 & 72.91 & 79.47 & 74.73 & 60.03 & 64.14 \\
SIFA v2     & 5.15  & 19.10 & \textbf{4.97}  & 47.55 & 13.75 & 43.72 & 19.20 & 21.92 \\
MAPSeg      & 17.08 & 14.11 & 9.17  & 45.88 & 37.68 & 23.49 & 36.60 & 26.29 \\
UMMKD       & \textbf{3.54}  & \textbf{6.08}  & 4.33  & 60.84 & 29.84 & 21.98 & 35.28 & 23.13 \\
\textbf{UnDA (Ours)} & 6.48 & 16.13 & 9.11 & \textbf{24.27} & \textbf{7.57} & \textbf{19.29} & \textbf{12.54} & \textbf{13.94} \\
\hline
\end{tabular}
\end{table*}

UnDA achieves the second-highest mean Dice (82.71\%), behind UMMKD (85.99\%) but ahead of
SIFA v2 (71.33\%), DAR-UNet (63.74\%), MAPSeg (61.35\%), and SE\_ASA (51.23\%). UnDA attains
the best Dice among all compared methods on RA (82.55) and is competitive on LA (89.58) and
AA (85.66). While UMMKD achieves higher Dice on LV, RV, and PA, UnDA shows more balanced
performance across structures, with no single structure falling as low as SE\_ASA's RV score
(10.02) or MAPSeg's PA score (21.48).

In terms of boundary accuracy, UnDA achieves the best mean HD95 (13.94\,mm) among all
compared methods, improving on UMMKD (23.13\,mm) by 9.19\,mm and on SIFA v2 (21.92\,mm) by
7.98\,mm, despite UMMKD and SIFA v2 posting lower HD95 on individual structures such as LV
Myo and LV. This suggests that although some baselines achieve marginally higher Dice on
select structures, UnDA produces more consistent and reliable boundary delineation overall
— an important property for clinical deployment where large localized boundary errors are
more costly than small uniform ones.

\section{Ablation Study}

The MM-WHS results in Table~\ref{tab:ablation} serve as our ablation study, isolating the contribution of each component. Adding $\mathcal{L}_{\text{pNCE}}$ alone to the no-alignment baseline yields inconsistent improvements: gains on RA ($+3.99$) and PA ($+24.66$) are offset by regressions on LV Myo ($-1.51$) and LA ($-1.80$), suggesting prototype-level alignment alone is insufficient to bridge large modality gaps uniformly. Plain OT without uncertainty weighting performs worse than the no-alignment baseline overall ($-0.92$ mean Dice), with a notable drop on PA ($-6.56$), 
indicating that unweighted transport is misled by noisy source samples for rare, small structures. UCT-OT recovers this by suppressing uncertain anchor tokens during transport, yielding substantial gains on LV ($+5.81$) and PA ($+28.84$) over the no-alignment baseline. Combining 
UCT-OT with $\mathcal{L}_{\text{pNCE}}$ in the full model further improves RA ($+7.31$ over UCT-OT alone) and PA ($+1.53$), confirming that the two objectives are complementary: $\mathcal{L}_{\text{UCT-OT}}$ drives uncertainty-aware global distribution alignment while $\mathcal{L}_{\text{pNCE}}$ enforces per-class prototype discriminability 
for structures with high inter-modal appearance variability.

\begin{table}[!ht]
\centering
\caption{Ablation study on MM-WHS (MRI $\to$ CT). Dice (\%, $\uparrow$) and HD95 (mm, $\downarrow$) per cardiac structure and 7-structure mean. \textbf{Bold} indicates best result per column.}
\resizebox{\textwidth}{!}{%
\label{tab:ablation}
\begin{tabular}{p{3.5cm}cccccccc}
\hline
Method & LV Myo & LA & LV & RA & RV & AA & PA & 7-Mean \\
\hline
\multicolumn{9}{c}{\textit{Dice Score (\%) $\uparrow$}} \\
\hline
Anchor: MR & 76.21 & 72.71 & 90.14 & 75.87 & 85.14 & 54.26 & 56.43 & 72.96 \\
Target: CT & 79.83 & 76.66 & 84.03 & 65.00 & 68.85 & 65.80 & 60.94 & 71.59 \\
MR $\to$ CT & 83.78 & \textbf{89.58} & 84.36 & 73.45 & 74.06 & 85.63 & 47.16 & 76.86 \\
\quad + ProtoNCE & 82.27 & 87.78 & 83.81 & 77.44 & 74.74 & 85.73 & 71.82 & 80.51 \\
\quad + OT & 83.80 & 86.33 & 85.14 & 74.95 & 72.77 & \textbf{87.97} & 40.60 & 75.94 \\
\quad + UCT-OT & 85.19 & 86.88 & \textbf{90.17} & 75.24 & 79.96 & 83.08 & 76.00 & 82.36 \\
\quad + UCT-OT + ProtoNCE (UnDA) & \textbf{84.85} & 85.53 & 86.74 & \textbf{82.55} & \textbf{76.10} & 85.66 & \textbf{77.53} & \textbf{82.71} \\
\hline
\multicolumn{9}{c}{\textit{HD95 (mm) $\downarrow$}} \\
\hline
Anchor: MR & 4.32 & 32.82 & \textbf{2.75} & \textbf{7.18} & 6.26 & 25.53 & 18.03 & 13.84 \\
Target: CT & 7.49 & 29.49 & 10.17 & 42.35 & 18.00 & 13.05 & 29.29 & 21.41 \\
MR $\to$ CT & 3.42 & \textbf{2.61} & 11.17 & 50.25 & 7.05 & 8.97 & 11.59 & 13.58 \\
\quad + ProtoNCE & 4.37 & 4.87 & 11.48 & 43.30 & 6.94 & 9.91 & 37.47 & 16.91 \\
\quad + OT & 7.93 & 26.66 & 11.60 & 40.61 & 15.57 & \textbf{6.35} & 21.68 & 18.63 \\
\quad + UCT-OT & \textbf{2.28} & 5.63 & 9.11 & 50.94 & \textbf{6.28} & 9.17 & 36.43 & 17.12 \\
\quad + UCT-OT + ProtoNCE (UnDA) & 6.48 & 16.13 & 11.32 & 24.27 & 7.57 & 19.29 & \textbf{12.54} & \textbf{13.94} \\
\hline
\end{tabular}
}
\label{tab:ablation}
\end{table}

\section{Conclusion}

We introduce UnDA, an anchor-guided framework for unpaired cross-modal segmentation that transfers structural knowledge via class-token alignment. Combining prototype-guided learning with uncertainty-weighted optimal transport enables stable feature alignment without paired data or spatial correspondence. A two-stage strategy lets the anchor model provide structural guidance while keeping inference efficient—only the target model is deployed. Experiments show consistent improvements across cross-modal medical segmentation tasks, validating uncertainty-aware token-level alignment.

\end{document}